%% file: main.tex
\documentclass[letterpaper, 10 pt, conference]{ieeeconf}  
\IEEEoverridecommandlockouts                              
\usepackage{cite}
\usepackage{amsmath,amssymb,amsfonts}
\usepackage{algorithmic}
\usepackage{textcomp}
\usepackage{xcolor}
\usepackage{multirow}

\usepackage{float}
\usepackage[caption = false]{subfig}
\usepackage[final]{graphicx}

\usepackage{mathtools}
\usepackage{dsfont}
\usepackage{nicefrac} 
\usepackage{csquotes}
\let\labelindent\relax
\usepackage[inline]{enumitem}

\usepackage{adjustbox}

\usepackage{algorithm}
\usepackage{algorithmic}

\usepackage{cleveref}
\usepackage{subfiles}

\title{\LARGE \bf
Towards Optimal Energy Management Strategy for Hybrid Electric Vehicle with Reinforcement Learning
}

\author{Xinyang~Wu$^{1}$, Elisabeth~Wedernikow$^{1}$,
Christof~Nitsche$^{1}$, and Marco~F. Huber$^{1,2}$
\thanks{$^{1}$ X. Wu, E. Wedernikow, C. Nitsche, and M. F. Huber are with the Department Cyber Cognitive Intelligence (CCI), Fraunhofer IPA. \{\texttt{xinyang.wu}, \texttt{elisabeth.wedernikow},
 \texttt{christof.nitsche}\} \texttt{@ipa.fraunhofer.de}
}%
\thanks{$^{2}$ M. F. Huber is also with the Institute of Industrial Manufacturing and Management IFF, University of Stuttgart. \texttt{marco.huber@ieee.org}
}%
}

\begin{document}

\maketitle
\thispagestyle{empty}
\pagestyle{empty}

\begin{abstract}

In recent years, the development of Artificial Intelligence (AI) has shown tremendous potential in diverse areas. Among them, reinforcement learning (RL) has proven to be an effective solution for learning intelligent control strategies. As an inevitable trend for mitigating climate change, hybrid electric vehicles (HEVs) rely on efficient energy management strategies (EMS) to minimize energy consumption. Many researchers have employed RL to learn optimal EMS for specific vehicle models. However, most of these models tend to be complex and proprietary, making them unsuitable for broad applicability. This paper presents a novel framework, in which we implement and integrate RL-based EMS with the open-source vehicle simulation tool called FASTSim. The learned RL-based EMSs are evaluated on various vehicle models using different test drive cycles and prove to be effective in improving energy efficiency.
\end{abstract}

\section{Introduction}
\label{sec:1}

Nowadays, fuel-powered vehicles cause widespread social concerns due to climate change and limited fossil fuel supply \cite{amjad2010review, hannan2014hybrid, ali2018towards}. The electrification of the automobile is a promising solution to overcome these problems. However, the development of electric vehicles encounters technical difficulties \cite{zhang2015comprehensive}. As a compromise, hybrid electric vehicles (HEVs) have emerged as a promising technology for reducing fuel consumption and emissions within the current infrastructure \cite{martinez2016energy}, \cite{lian2020rule}, which offer a balance of environmental benefits, fuel economy, and driving performance. For example, in the last ten years, the proportion of the newly registered HEVs in all kinds of vehicles has risen from less than $0.1\,\%$ to $31.2\,\%$ in the German market~\cite{pkw-kba}. 

HEVs generally have multiple sources to power the drivetrain. Therefore some sort of energy management is required to manage the cooperation of the individual power components. Some common strategies include:
\begin{enumerate*}[label=(\arabic*)]
    \item \emph{Charge Depleting (CD)} strategy, which uses the electric motor to power the vehicle as much as possible, and only switches to the internal combustion engine (ICE) when the battery's charge is depleted,
    \item \emph{Charge Sustaining (CS)} strategy, which maintains a constant state of charge (SOC) in the battery by adjusting the power split between the electric motor and the ICE,
    \item \emph{Power-Split} strategy, which uses both the electric motor and ICE to power the vehicle simultaneously. The power split between the two is adjusted to achieve optimal fuel efficiency.
\end{enumerate*}


In this work, we focus on the Power-Split strategy to achieve optimal efficiency for different HEVs, also referred to as the energy management strategy (EMS) in HEVs. Many automobile manufacturers have developed their own specific software for optimizing the EMS for their vehicles, and much of the research lacks a common platform as the baseline. This work aims to implement a framework to optimize the EMS of various HEVs with reinforcement learning (RL) in an open-source vehicle powertrain simulation tool, namely the Future Automotive Systems Technology Simulator (FASTSim) \cite{brooker2015fastsim}. 


Compared to the state-of-the-art (SotA), the contributions of this work are as follows:

\begin{enumerate}[label=(\arabic*)]
    \item We provide an open-source solution that leverages RL algorithms to learn optimal EMS in different driving situations. We re-programmed FASTSim, originally designed with a rule-based strategy, to be compatible with RL-based strategies. This is especially useful for researchers in the RL community.
    \item Most SotA methods depend on Matlab or proprietary software for building specific vehicle models. In contrast, we offer generalized interfaces for various vehicle models and different driving cycles. 
    \item Many SotA methods hard-code boundary constraints, such as speed requirements, whereas we encode constraints as parts of the reward function and let the RL-agent learn to obey them during exploration.
\end{enumerate}


\section{Related Work}
\label{sec:2}

\subsection{Energy Management Strategy}

The EMS in HEVs is often realized using rule-based approaches, which include deterministic rule-based and fuzzy rule-based methods \cite{hofman2007rule, li2011energy}. Experienced engineers must carefully design the rules to achieve the desired behavior. When designed correctly, the rule-based approach provides energy management with real-time capabilities and high accountability. Since the rules are hard-coded, the model, however, has limited flexibility \cite{ali2018towards, lian2020rule} and cannot fully exploit the potential fuel savings \cite{gonder2008route}. 

Another popular approach is the optimization-based method, such as model predictive control \cite{camacho2013model} or dynamic programming \cite{bellman1966dynamic, peng2017rule, panday2014review}. In such methods, a mathematical model of the HEV system is used to predict the vehicle's energy needs and determine the optimal power split between the electric motor and the ICE. The optimization algorithm takes into account factors such as the vehicle's current speed, the SOC of the battery, the engine load, and the driver's requested power. The optimization goal is to minimize the fuel consumption of the HEV while meeting the driver's power demand. To some extent, these methods improve the real-time performance and fuel economy of EMS \cite{onori2010adaptive, jinquan2019novel, lian2020rule}. However, such methods require complex mathematical models and high computational resources.

\begin{figure}[t]
    \centering
    \includegraphics[width=0.4\textwidth]{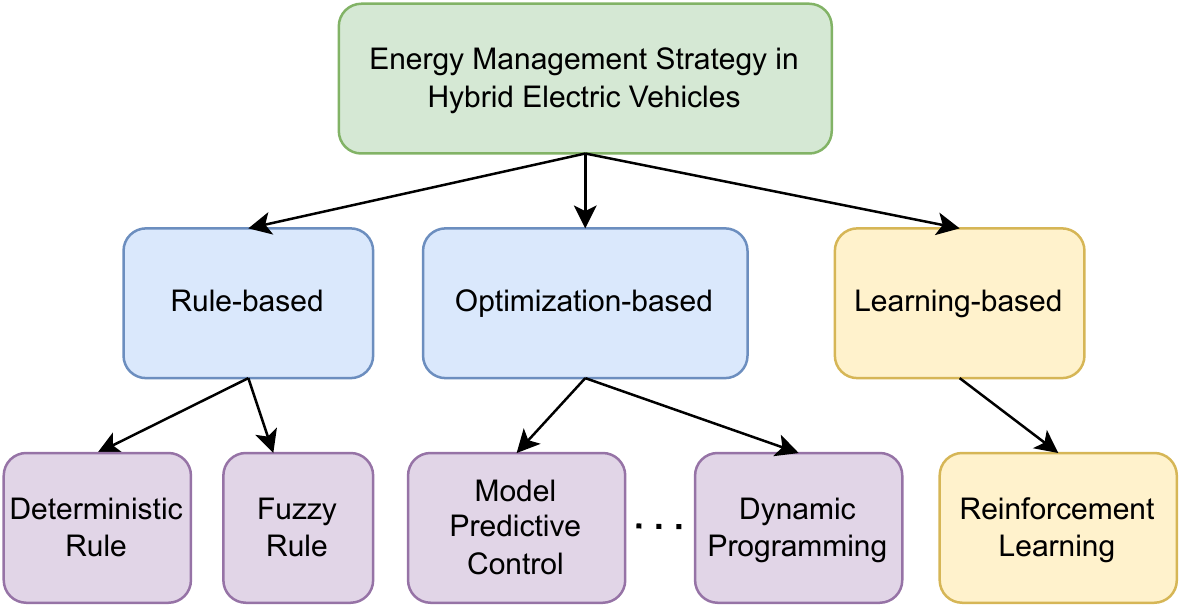}
    \caption{State of the art for the optimization of EMS in HEVs \cite{panday2014review, rudolf2021toward}.}
    \label{fig:sota}
\end{figure}

Recently, learning-based methods have been suggested to learn an appropriate EMS automatically. Especially RL-based methods showed promising results, which are more flexible than the rule-based approach and are also real-time capable \cite{lian2020rule}. Several works have already been conducted to examine the benefits and difficulties of using RL for energy management in HEVs. RL algorithms, including deep deterministic policy gradient (DDPG) \cite{tan2019energy, wu2019deep} and a variety of Q-learning algorithms have been tested to solve different energy management tasks \cite{liu2015reinforcement, lian2020rule}. Figure~\ref{fig:sota} shows an overview of the three different kinds of methods for optimizing the operational strategy of energy management in HEVs.

\subsection{Reinforcement Learning}

RL is a specific machine learning approach where an agent learns to make decisions by interacting with an environment and receiving feedback in the form of rewards \cite{sutton2018reinforcement}. The goal of the agent is to learn a policy, which is a mapping from states of the environment to actions. The goal is to maximizes the cumulative reward over time. RL has been applied to various control problems, such as robotics, game-playing, and autonomous vehicles. It has also been used to optimize the EMS in HEVs to achieve better fuel efficiency, such as \cite{zou2016reinforcement, wu2018continuous, qi2019deep, zhang2019energy, lian2020rule}.

Many different algorithms can be used for RL, such as Q-learning \cite{watkins1992q}, SARSA \cite{rummery1994line}, actor-critic RL \cite{konda1999actor}, and policy gradient \cite{sutton1999policy}. The choice of the algorithm will depend on the specific problem and the type of environment. In this paper, we use DDPG \cite{lillicrap2015continuous}, which is based on the actor-critic structure and utilizes deep neural networks (NNs) to generalize for continuous state and action spaces. The RL algorithm aims to learn a policy, which maps the vehicle's states to actions that maximize the cumulative reward over time.

\subsection{Prioritized Experience Replay}

In most model-free and off-policy RL settings, the trajectories experienced by the RL-agent are usually saved in the replay buffer, and trajectories will be uniformly sampled and learned. However, some trajectories may have more information than others, and thus they should be more frequently chosen to be learned. The idea is similar to importance sampling. In RL, the importance of different trajectories can be quantified by various metrics, and \cite{schaul2015prioritized} proposed prioritized experience replay (PER), where the trajectories are sampled based on the temporal difference error (TD-Error).

\subsection{Monte Carlo Dropout}

Dropout is a technique that has been proposed to improve the generalization and to suppress overfitting of deep NNs \cite{hinton2012improving, srivastava2014dropout, baldi2013understanding} by introducing a form of model uncertainty into the predictions made by deep NNs. By using dropout prior to of each layer, Monte Carlo dropout (MC dropout) \cite{gal2016dropout} proposes to train the deep NN to approximate the underlying Gaussian Process \cite{damianou2013deep}. MC dropout has been proven to improve the generalization and prediction performance further.

\subsection{FASTSim}

Different from specific vehicle models or proprietary software, FASTSim is designed to be open-source, computationally lightweight, accurate, and scalable, offered by the National Renewable Energy Laboratory (NREL), USA \cite{brooker2015fastsim, gonder2018future}. It provides Python implementations and a relatively simple approach to compare different vehicle powertrains on vehicle efficiency, performance, and battery life. Users can either select vehicles already predefined in FASTSim or model various vehicles by different parameters, such as vehicle weights, battery capacity, engine powers, and so on. Additionally, various driving cycles can be imported to test the vehicle model, such as the Urban Dynamometer Driving Schedule (UDDS)~\cite{EPA} or the Worldwide Harmonised Light Vehicle Test Procedure (WLTP)~\cite{tsiakmakis2017nedc}. Therefore, variations of the vehicle or powertrain can be assessed under different driving conditions. 

\begin{figure*}[t]
\centering
\subfloat[Components of FASTSim.]{\includegraphics[width=0.65\textwidth]{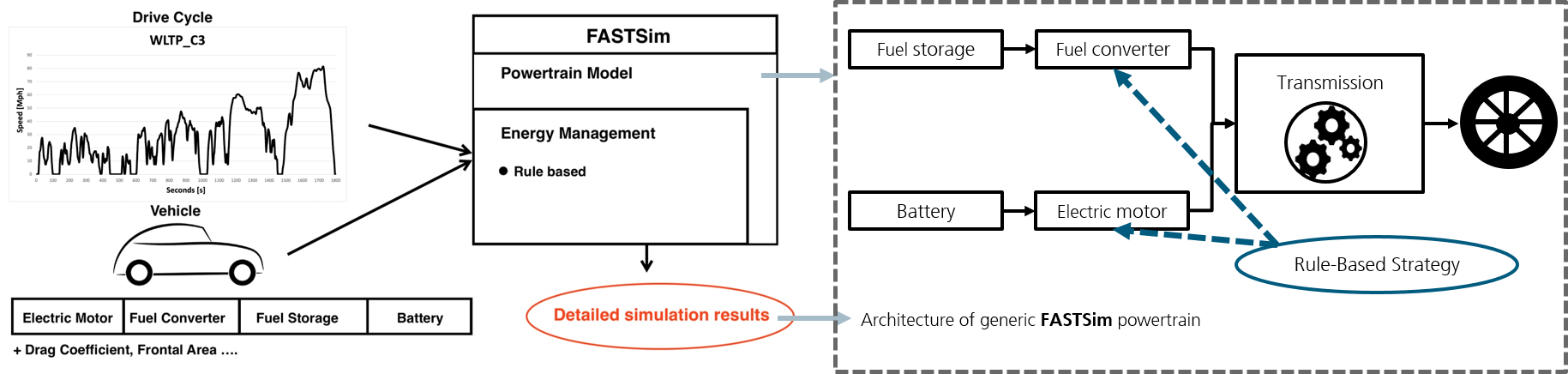}} 
\hspace{1.0cm}
\subfloat[Example of the simulation result for the driving speed.]{\includegraphics[width=0.26\textwidth]{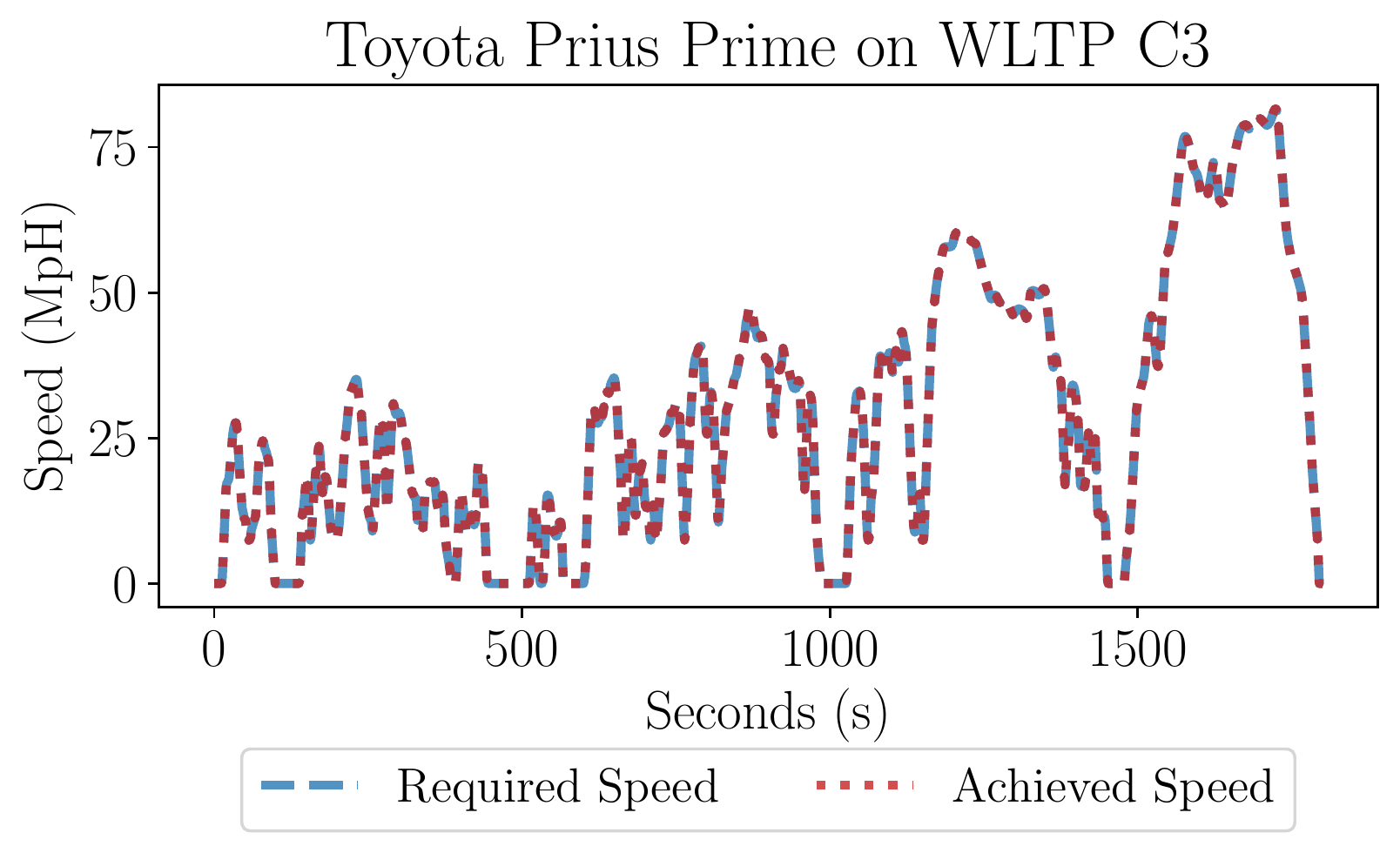}}

\caption{(a) Important components of FASTSim for the simulation. Vehicle parameters and a driving cycle are required as the input, and FASTSim enables further customization of the vehicle models or self-designed driving cycles. The powertrain model in FASTSim will calculate the achieved speeds, accelerations, motors' output powers, and energy consumption of the vehicle for completing the driving cycle. The power-split between ICE and the electric motor is determined by the rule-based strategy, which is designed by FASTSim for an accurate simulation of real-world situations. (b) shows the example of one simulation result on the WLTP-C3 driving cycle of the Toyota Prius Prime.}
\label{fig:fastsim-components-example}
\end{figure*}

FASTSim simulates the vehicle and its components through speed-vs-time drive cycles. At each timestep, FASTSim accounts for drag, acceleration, ascent, rolling resistance, regenerative braking, each powertrain component’s efficiency, and power limits \cite{brooker2015fastsim}. The vehicle models are simplified to some extent. Therefore, a scalable and generalized simulation of different kinds of vehicles becomes possible. Figure \ref{fig:fastsim-components-example} gives an overview of the components in FASTSim.

\section{Deep RL Framework with FASTSim}
\label{sec:3}

In this work, we propose a framework for training RL-agents to learn driving strategies for various HEVs using FASTSim. The proposed framework provides a systematic approach for the training, simulation, and validation of the RL-based driving strategies.

\begin{figure*}[t]
    \centering
    \includegraphics[width=0.75\textwidth]{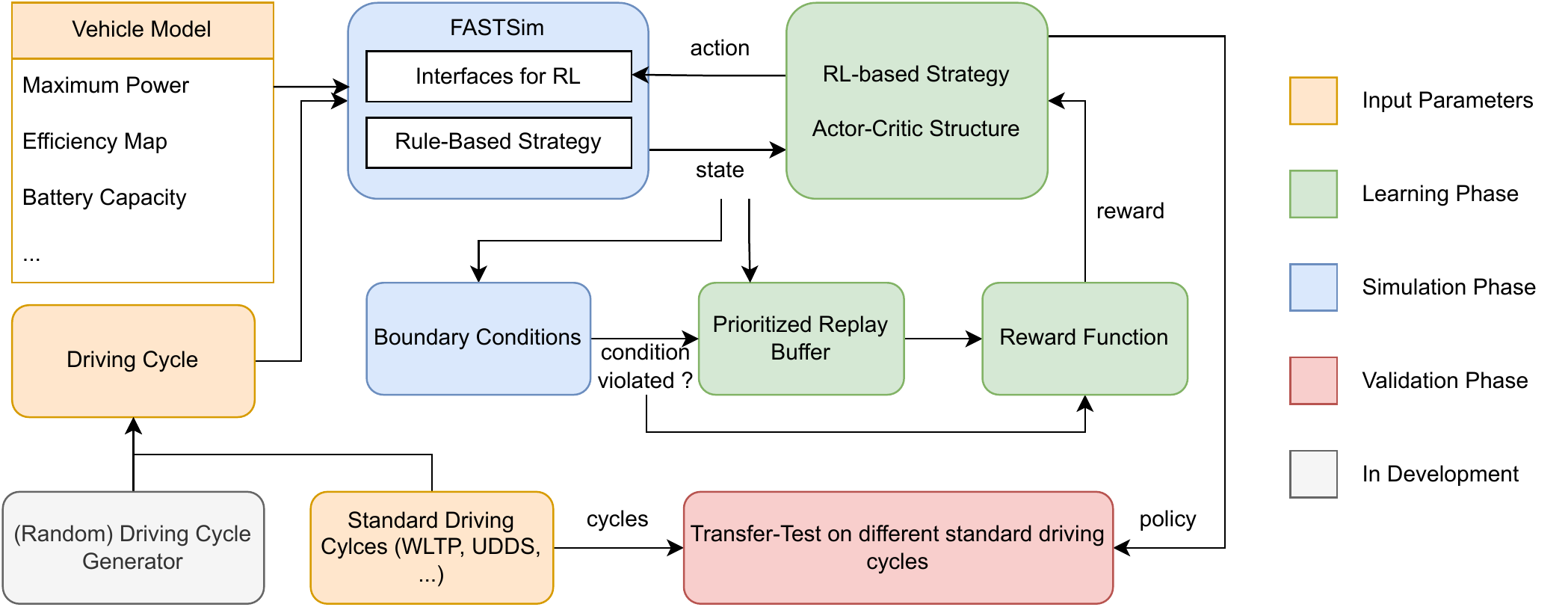}
    \caption{The proposed framework for training RL-based driving strategies for HEVs using FASTSim. The combination of FASTSim in this framework allows for evaluating strategies for different HEVs under a diverse set of driving conditions, resulting in more generalizable driving strategies.}
    \label{fig: framework}
\end{figure*}

The framework consists of four main blocks: input parameters, learning phase, simulation phase, and validation phase. The \emph{input parameters} allow for the customization of the framework to different driving scenarios and HEVs. In the \emph{simulation phase} and \emph{learning phase} blocks, the agent interacts with the simulated environment and the agent's policy is updated based on trials and rewards, respectively. Finally, the \emph{validation phase} evaluates the performance of the learned strategies on various driving cycles and validates the transferability of the RL-agent. Figure~\ref{fig: framework} shows the overview of the framework.

\subsection{Input Parameters}

\paragraph{Vehicle Model}

The vehicle is represented by a set of parameters in FASTSim. Top-level parameters like frontal area and drag coefficient describe the physical properties of the vehicle as a complete unit. These parameters play a prominent role in road load equations. The road load equations are implemented in FASTSim to estimate the power required for the vehicle to meet the drive cycle. Low-level parameters represent the powertrain components of the vehicle, the ICE, the electric motor, fuel storage, and the battery. The parameters of the transmission components are pre-defined values that describe the properties of the component and constrain the behavior of the transmission. Table~\ref{table:param-fastsim} lists the main parameters of the electric motor. 

\begin{table}
\caption{Low-level parameters for electric motors in Fastsim.}
\label{table:param-fastsim}
\begin{center}
\begin{tabular}{|l||l|}
\hline
\begin{tabular}{@{}l@{}} Maximum Power / kW \\ Time to Full Power / s\end{tabular} & 
\begin{tabular}{@{}l@{}} Limit the acceleration performance and  \\ maximum speed of the electric motor \end{tabular} 
\\
\hline
\begin{tabular}{@{}l@{}} Bass Mass / kg \\ Specific Power / 
 kg/Kw\end{tabular} &
\begin{tabular}{@{}l@{}} Estimate and scale the mass of the electric \\ motor based on power\end{tabular}
\\
\hline
\begin{tabular}{@{}l@{}} Efficiency Curve \end{tabular} & 
\begin{tabular}{@{}l@{}} Efficiency at different power output \\ percentages \end{tabular}
\\
\hline
\end{tabular}
\end{center}
\end{table}

\paragraph{Driving Cycle}

A driving cycle, also known as a drive cycle or test cycle, is a standardized driving pattern used to evaluate the performance of a vehicle. The driving cycle consists of a series of speed and acceleration commands that simulate a specific driving scenario, such as city or highway driving. In FASTSim, a driving cycle is used to simulate the vehicle's dynamic behavior, fuel economy, and other performance characteristics under different driving conditions. This allows FASTSim to evaluate the performance of various HEVs and the effectiveness of different EMS under the specific driving conditions represented by the driving cycle. Many different driving cycles have been developed for use in vehicle testing, including the UDDS, WLTP, Highway Fuel Economy Test (HWFET)~\cite{EPA}, New European Driving Cycle (NEDC)~\cite{tsiakmakis2017nedc}, and the US06 Supplemental Federal Test Procedure (US06)~\cite{EPA}.

\paragraph{Driving Cycle Generator}

An RL-based strategy trained on a specific driving cycle may exhibit overfitting, leading to sub-optimal performance when applied to other driving situations or behaviors. To mitigate this issue, a random driving cycle generator can be implemented to increase the diversity of the training dataset. This can be achieved by incorporating noise, concatenating, or cropping standard driving cycles. The incorporation of a diverse set of driving cycles in the training dataset can lead to a more generalizable RL-based strategy, thus improving its performance in various driving conditions.

\subsection{Learning Phase and Simulation Phase}

In the \emph{learning phase}, the agent interacts with the simulated environment and the agent's policy is updated based on the observed rewards as feedback. During this phase, the agent learns from its own experiences and improves its decision-making over time. In the \emph{simulation phase}, FASTSim allows the agent to explore different driving scenarios and conditions and learn a driving strategy robust to diverse operating conditions. 

\paragraph{Reward Function}

In our approach, the reward function is designed to encourage the agent to learn a driving strategy that maximizes the energy efficiency of the vehicle while meeting the driving constraints, which assigns negative rewards for actions that result in low energy efficiency or violate the driving constraints. The state, action, and reward are defined as

\begin{align}
    \mathrm{state}~ =& ~ \{ \mathrm{SOC}, s_\mathrm{cycle}, a_\mathrm{cycle}, s_\mathrm{achieved} \}~,  \\
    \mathrm{action}~ =& ~ p_\mathrm{ICE} / p_\mathrm{cycle}~,  \\
    \mathrm{reward}~ =& -\alpha_1 \cdot p_\mathrm{achieved}  ~ \nonumber \\
    & - \alpha_2 \cdot [\left | s_\mathrm{cycle} - s_\mathrm{achieved} \right | > 0]~ \nonumber \\
    & - \alpha_3 \cdot  [(\mathrm{SOC}_\mathrm{ref}-\mathrm{SOC}) > \beta ] ~, 
\label{eq:reward}
\end{align}
where $s_\mathrm{cycle}$, $a_\mathrm{cycle}$ and $p_\mathrm{cycle}$ represent the required speed, acceleration, and power of the given driving cycle, respectively, $s_\mathrm{achieved}$ and $p_\mathrm{achieved}$ are the achieved speed and actual output power of the vehicle model in simulation, respectively. Further, $p_\mathrm{ICE}$ is the output power of ICE, which refer to the power-splitting in HEVs. $\mathrm{SOC}_\mathrm{ref}$ means the reference SOC, which is a target value that represents the desired level of charge for the vehicle's battery. It ensures that the battery is operated within a safe and efficient range. $(\alpha_1, \alpha_2, \alpha_3)$ are the non-negative coefficients for balancing the fuel efficiency and boundary conditions, while $\beta > 0$ defines the threshold of the allowed difference between current $\mathrm{SOC}$ and $\mathrm{SOC}_\mathrm{ref}$. By using a well-designed reward function, the agent learns to take optimal actions that result in high energy efficiency, while keeping the $\mathrm{SOC}$ in a healthy working condition for the battery. We utilize Bayesian optimization to search for the best parameters. 

\paragraph{Boundary Conditions}

In our proposed framework 
for training RL-based strategies for HEVs utilizing FASTSim, certain boundary conditions have been implemented to ensure the validity and applicability of the obtained results. One important boundary condition is the correspondence between the speed of the given driving cycle and the vehicle model. This boundary condition is essential as any discrepancies between the speed of driving cycle $s_\mathrm{cycle}$ and the actual speed of the vehicle model $s_\mathrm{achieved}$ can lead to inaccuracies in the evaluation of the vehicle's energy consumption. To mitigate this and to encourage the RL-agent to focus on completing the driving cycle correctly, we assign a negative reward $ - \alpha_2$ to the agent, as long as $\left | s_\mathrm{cycle} - s_\mathrm{achieved} \right | > 0$ is evaluated as true.

\paragraph{Benchmark between Rule-based and RL-based Strategies}

\begin{figure}[t]
    \centering
    \includegraphics[width=0.45\textwidth]{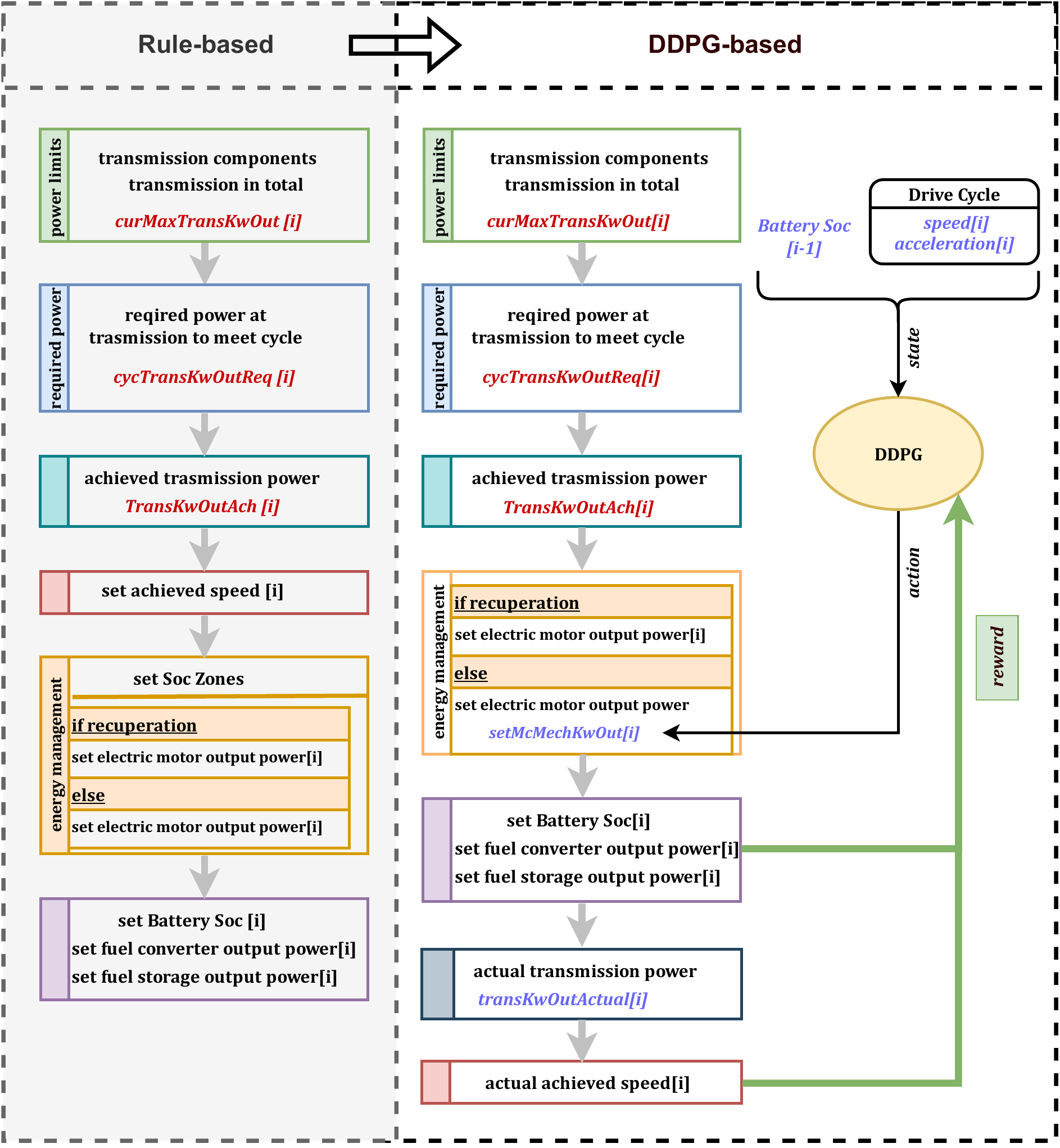}
    \caption{The comparison between the default rule-based strategy and the implementation of the RL-based strategy in the form of a UML diagram. Both strategies share low-level interfaces in FASTSim.}
    \label{fig:uml}
\end{figure}

To evaluate and compare the RL-based strategies against the default rule-based strategies provided by FASTSim, we implement the RL algorithm based on the interfaces of FASTSim. Figure~\ref{fig:uml} shows the comparison between the decision processes of both kinds of EMS in Unified Modeling Language (UML). The default rule-based strategies in FASTSim will first calculate the required output power for satisfying the driving cycle and then divide the power requirements between ICE and electric motor according to hard-coded rules. In contrast, the RL-based strategy will let the agent decide on the power-split itself. After that, the current $\mathrm{SOC}$ of the battery and the achieved speed $s_\mathrm{achieved}$ will be fed back to the RL algorithm, guiding itself for learning an optimized strategy.

\paragraph{Priorities for the Replay Buffer}

In PER, transitions are assigned a priority value that reflects their importance or information gain for learning. The priority value can be based on various factors, such as the TD-Error. Transitions with higher priority values are more likely to be replayed during the learning process. Here, we use PER in the replay buffer to further improve the sampling efficiency and stability of the RL algorithm by focusing the learning process on the most informative transitions. This can lead to faster convergence and better performance of the learned policy.

\subsection{Validation Phase}

In this framework, the validation phase plays a crucial role in ensuring the effectiveness and robustness of the learned RL-based strategies. The core component of the validation phase is the \emph{transfer test}, which involves testing the learned agent on different driving cycles. For example, we train the RL-agent on WLTP-C3 and evaluate it on the other driving cycles, such as NEDC, UDDS, and HWFET. The transfer test allows evaluating the agent's performance under different driving conditions and assessing its ability to adapt to new scenarios. This is especially important for real-world use, as driving conditions can vary greatly depending on the route, traffic, and weather conditions. To this end, the transfer test provides a robust and reliable evaluation of the learned RL-agent. It further allows ensuring that the agent is generalizable and effective under different scenarios.

\section{Experiments}
\label{sec:4}

\begin{figure*}[t]
    \centering
    \includegraphics[width=0.9\textwidth]{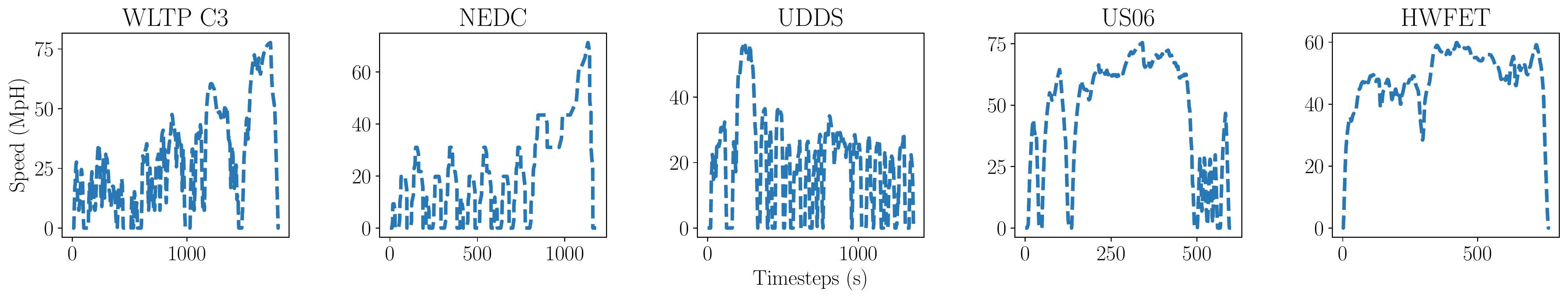}
    \vspace{-2mm}
    \caption{The five standard driving cycles used in our experiments. All RL-based strategies are trained on WLTP-C3 and tested on all the five cycles.}
    \label{fig:cycles}
\end{figure*}

In this section, we present the experimental results of our proposed framework, which were conducted under different driving cycles and with various HEVs. In the validation phase, we will show the transfer tests of the learned RL-agents on five different driving cycles, as shown in Figure~\ref{fig:cycles}. All the strategies are trained on WLTP-C3 and tested on the other cycles. We show the results on the following two HEVs, which both apply a power-split strategy (cf. Sec. \ref{sec:1}):
\paragraph{BMW i3 REx, 2016} a series plug-in hybrid vehicle with range extender, where the ICE only works with a generator to recharge the battery and is isolated from the axle. Its lithium-ion battery has a capacity of 94\,Ah (33\,kWh).
\paragraph{Toyota Prius Prime, 2017} a series-parallel plug-in hybrid vehicle that combines the concepts of series and parallel hybrid, in which the ICE not only recharges the battery with a generator but also drives the transaxle together with the electric motor in different modes. It has a smaller lithium-ion battery with a capacity of 8.8\,kWh.

\subsection{Results}
In our tests, most of the RL-agents are able to converge within 10 episodes for various HEVs on the WLTP-C3 driving cycle, thanks to the lightweight simulator and the PER buffer in our framework. In order to provide a comprehensive understanding of the learning process, we present the results of the RL-agent trained after 1 episode and after 10 episodes as a comparison. We also present the rule-based strategy's results as a benchmark. In all the experiments, we use NNs with two hidden layers with $100$ neurons each for the Actor. For the Critic, we utilize three hidden layers, incorporating $100$ neurons each for the first two layers and $50$ neurons for the third layer, with MC Dropout in front of each layer. In the reward function~\eqref{eq:reward}, we use $(1.5, 10, 0.1)$ for the three coefficients. The reason for the small value $\alpha_3 = 0.1$ is explained in Sec.~\ref{sec:limit}.

\begin{figure*}[t]
    \centering
    \includegraphics[width=0.95\textwidth]{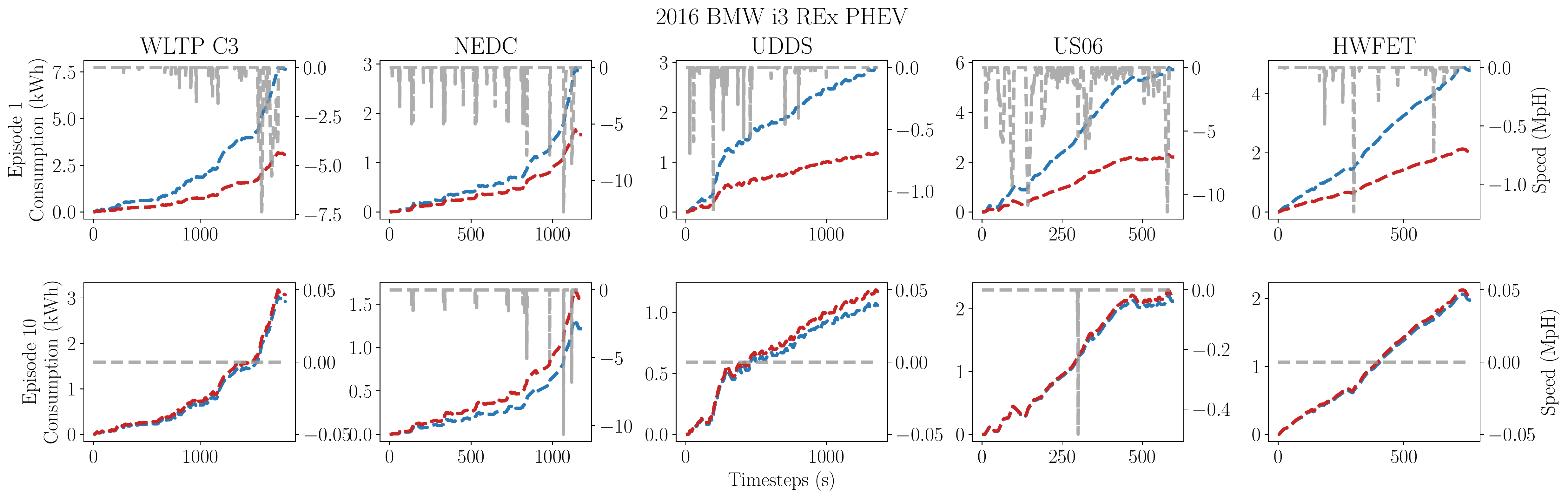}
    \includegraphics[width=0.95\textwidth]{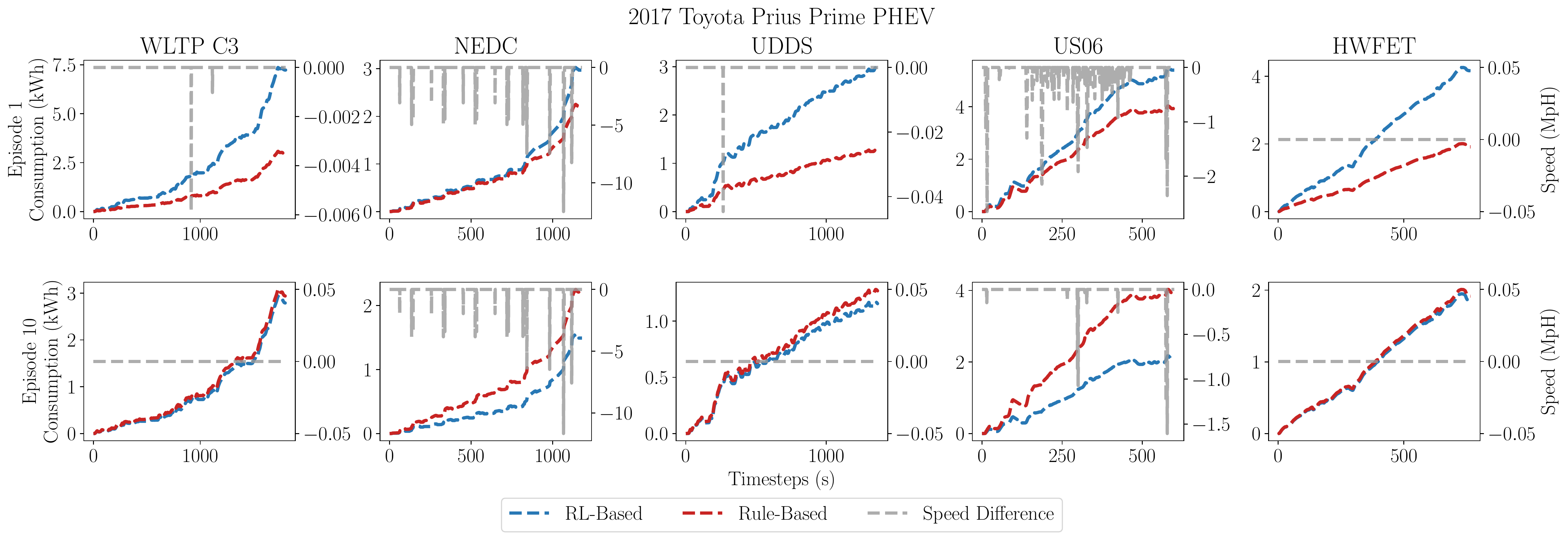}
    \caption{Transfer tests of the learned RL-agents for BMW i3 REx PHEV (the first two rows) and Toyota Prius Prime PHEV (the last two rows). Both are trained on WLTP-C3 and tested on all five driving cycles. The first and third rows show the evaluation of the RL-agent after training for one episode, while the second and fourth rows show the results after training for 10 episodes. Blue curves are the total energy consumption of RL-based strategies, while red curves are the consumption from the default rule-based strategies. Gray curves show the difference between the required speed by the given driving cycles and the achieved speed of the HEVs, operated by the learned RL-based strategies.}
    \label{fig:transfer}
\end{figure*}

\paragraph{Learning Process}
As shown in Fig.~\ref{fig:transfer}, the RL-agents can reduce their energy consumption for both HEVs and on all five driving cycles after training for 10 episodes. For example, the BMW i3 REx PHEV consumes $7.65$\,kWh on the WLTP-C3 driving cycle, with the RL-based strategy trained for merely one epoch. As a comparison, the total energy consumption reduces to $2.92$\,kWh after training for 10 episodes, which is even less than the $3.07$\,kWh of the rule-based strategy on the same driving cycle. Meanwhile, the learned strategy for the Toyota Prius Prime reduces its consumption from $7.23$\,kWh to $2.79$\,kWh after 10 episodes, while the rule-based strategy requires $2.94$\,kWh in total. 

When trained only for one episode, both HEVs fail to finish the WLTP-C3 driving cycle and show speed differences between their achieved speeds $s_\mathrm{achieved}$ and the required speeds $s_\mathrm{cycle}$ (cf. the first column of Fig.~\ref{fig:transfer}). Thanks to the added boundary condition in the reward function~\eqref{eq:reward}, the RL-agents learn to stricly follow the required speeds during the driving cycle when trained for ten episodes. Similar results are observed for many other HEV models, e.g., Chevrolet Volt, Ford C-MAX, and Hyundai Sonata, and prove the effectiveness of the learning process, as shown in Table~\ref{table:train_wltp}.

\begin{table}[t]
\centering
\caption{Comparison of the energy consumption (kWh).} 
\vspace{-2mm}
\label{table:train_wltp}
\tabcolsep=0.12cm
\scalebox{0.815}{\begin{tabular}{l|ccccc|ccccc}
\hline
&\multicolumn{5}{c}{rule-based} & \multicolumn{5}{c}{RL-based} \\\hline
Cycle & i3 & Prius & Volt & C-MAX & Sonata & i3 & Prius & Volt & C-MAX & Sonata \\\hline
WLTP\_C3 & $3.07$ & $2.94$ & $3.04$ & $3.78$ & $3.99$ & $2.92 $ & $2.79$ & $2.89$ & $3.47$ & $3.29$\\
UDDS & $1.17$ & $1.27$ & $ 1.21 $ & $1.40$ & $1.46$ & $1.06$ & $1.16$ & $1.09$ & $1.30$ & $1.34$\\
HWFET & $2.04$ &  $1.92$ & $2.00 $ & $ 2.41 $ & $2.24$ & $1.98$ & $1.85$ & $1.94$ & $2.34$ & $2.17$\\
\hline
\end{tabular}}
\end{table}

\paragraph{Transferability}
To validate the transferability and generalization of the learned RL-based strategies on different driving conditions, we evaluate the performance of the RL-agents on the other four driving cycles. As shown in Figure~\ref{fig:transfer}, both agents can ensure no speed difference and correctly follow the cycles on UDDS and HWFET after 10 episodes, similar as on WLTP-C3; however, they fail to satisfy the speed requirements on the NEDC and US06 driving cycles. A possible reason is that US06 contains more challenging driving situations, such as higher average speed and more aggressive acceleration, while NEDC is designed to have more urban driving phases (66\,\%) compared to WLTP-C3 (52\,\%). Compared to the rule-based strategies, both of the RL-agents achieve to reach relatively lower or similar total energy consumption after 10 episodes on WLTP-C3, UDDS and HWFET driving cycles, which proves the general applicability of our framework in different conditions. 

\subsection{Limitations}
\label{sec:limit}

In an effort to integrate RL algorithms with a suitable simulation environment for HEVs, we chose FASTSim as the basis of the framework and implemented several well-known RL algorithms. However, there are two limitations that we will focus on in future work.

\paragraph{Efficiency Map} instead of a complete efficiency map including different efficiency factors based on torque and speed of the engine, FASTSim adopts a simplified efficiency curve, where the efficiency rates depend merely on the output power of the engine. With such simplification, FASTSim provides a fast and lightweight simulation tool. However, it may lead to inaccurate simulation results of the learned RL-based strategies, which are dependent on the efficiency factors of different vehicle models.

\paragraph{Trade-off with Battery Lifetime} in the reward function~\eqref{eq:reward}, we include the factor $\alpha_3$ to balance the overload of the battery with energy efficiency. Theoretically, $\mathrm{SOC_\mathrm{ref}}$ represents the optimal working conditions of the battery. However, such parameters are lacking in FASTSim. In our experiments, we assume $65\%$ as the $\mathrm{SOC_\mathrm{ref}}$ for all HEVs and use a rather small factor $\alpha_3$ to assign negative rewards if the current $\mathrm{SOC}$ relative to $\mathrm{SOC_\mathrm{ref}}$ is smaller than threshold $\beta$. A more sophisticated formulation and the fine-tuning of $\alpha_3$ needs to be considered when more parameters about the batteries of different HEVs are available.

\section{Conclusion}
\label{sec:5}

We provide a systematic framework for the training, simulation, and validation of the RL-based driving strategies for various HEVs under different driving conditions. Several boundary conditions and transfer tests are incorporated to ensure the validity and applicability of the learned RL-agents in real-world scenarios. The experimental results show the potential of using RL for improving the performance and energy efficiencies of HEVs. In future work, we will focus on integrating the proposed framework and evaluating the learned strategies in a real-world vehicle and bridge the gap between simulation and reality.


\section*{Acknowledgment}
This work was partially supported by the Baden-W\"urttemberg Ministry of Economic Affairs, Labor, and Tourism within the KI-Fortschrittszentrum ``Lernende Systeme and Kognitive Robotik'' under Grant No. 036-140100.


\input{main.bbl}

\end{document}